# Fuzzy Rule based Intelligent Cardiovascular Disease Prediction using Complex Event Processing

Shashi Shekhar Kumar, Anurag Harsh, Ritesh Chandra, Sonali Agarwal

rsi2020502@iiita.ac.in*[a], iib2020016@iiita.ac.in[b] ,rsi2022001@iiita.ac.in[a] , sonali@iiita.ac.in[d]

Department of Information Technology

India Institute of Information Technology Allahabad, Prayagraj, India

**Abstract**

Cardiovascular disease (CVDs) is a rapidly rising global concern due to unhealthy diets, lack of physical activity, and other factors. According to the World Health Organization (WHO), primary risk factors include elevated blood pressure, glucose, blood lipids, and obesity. Recent research has focused on accurate and timely disease prediction to reduce risk and fatalities, often relying on predictive models trained on large datasets, which require intensive training. An intelligent system for CVDs patients could greatly assist in making informed decisions by effectively analyzing health parameters. Complex Event Processing (CEP) has emerged as a valuable method for solving real-time challenges by aggregating patterns of interest and their causes and effects on end users. In this work, we propose a fuzzy rule-based system for monitoring clinical data to provide real-time decision support. We designed fuzzy rules based on clinical and WHO standards to ensure accurate predictions. Our integrated approach uses Apache Kafka and Spark for data streaming, and the Siddhi CEP engine for event processing. Additionally, we pass numerous cardiovascular disease-related parameters through CEP engines to ensure fast and reliable prediction decisions. To validate the effectiveness of our approach, we simulated real-time, unseen data to predict cardiovascular disease. Using synthetic data (1000 samples), we categorized it into "Very Low Risk, Low Risk, Medium Risk, High Risk, and Very High Risk." Validation results showed that 20% of samples were categorized as very low risk, 15-45% as low risk, 35-65% as medium risk, 55-85% as high risk, and 75% as very high risk.

Table 1 List of Abbreviations

| Term used | Full Name |
|---|---|
| CVDs | Cardiovascular Disease |
| WHO | World Health Organization |
| CEP | Complex Event Processing |
| NCDs | Noncommunicable Diseases |
| AUROC | Area under the ROC curve |
| LSTM | Long Short Term Memory |
| CNN | Convolution Neural Network |
| ECG | Electrocardiogram |
| FCL | Fuzzy Control Language |
| DBP | Diastolic Blood Pressure |
| RNN | Recurrent Neural Network |
| SBP | Systolic Blood Pressure |
| SVR | Support Vector Regression |
| ACC | American College of Cardiology |

Table 1 shows the abbreviations used throughout this manuscript.

# 1 Introduction

Health is one of the most challenging issues for humans, necessitating immediate attention, precautionary measures, and expert consultation to mitigate the impact of any disease as well as reduce fatalities. CVDs is a major health issue that can be life-threatening if not treated properly within the timeline. However, adequate and timely measures of human parameters can reduce risk significantly.

According to the WHO recent reports[1] approximately 17.9 million people died from CVDs in 2019, representing 32% of all global deaths. Out of these casualties, 85% were due to heart attacks and strokes. In 2019, CVDs caused 38% of the 17 million premature deaths (under the age of 70) due to noncommunicable diseases.

CVDs can be prevented[2] by addressing behavioral risk factors such as usage of tobacco, unhealthy diet, obesity, physical inactivity, and the harmful use of alcohol. It is important to detect cardiovascular disease as early as possible so that precautionary measures can be provided to patients. Subsequently, the clinical data of human plays an important role in identifying the risk of CVDs.

Similarly, a report published by the ACC[3] in the context of India suggests that the mortality rate from NCDs accounted for 65% of total deaths in India in 2019, of which more than 25% could be attributed to cardiovascular diseases and related risk factors, which is an alarming situation.

---

[1] https://www.who.int/news-room/fact-sheets/detail/cardiovascular-diseases-(cvds)?gad_source=1&gclid=CjwKCAjwt-OwBhBnEiwAgwzrUpeh5SoPVI3sMvMeymxzCvKPDhN8MFGerLuRwP-HK4Nx6vf3t2N1ABoCvWcQAvD_BwE

[2] https://iris.who.int/bitstream/handle/10665/43787/9789241547260_eng.pdf

[3] https://www.acc.org//-/media/Non-Clinical/Files-PDFs-Excel-MS-Word-etc/2023/05/I23115-Advancing-CV-Health-in-India-Roundtable-Report.pdf

Automating the prediction of cardiovascular risk requires a lot of data to be analyzed and expert consultation is needed to timely diagnose the disease. On the contrary, delays in important parameter processing may increase risk further and may lead to serious health consequences [1]. Therefore it is essential to design such systems that work in real time with predefined set of rules to timely decision support for cardiovascular disease.

In developing such a framework, leveraging technological aspects becomes an important part, as it helps us make effective decisions for all. Exploring these tools to process large amounts of health parameters and detect a useful pattern, CEP [2] promises to be one of the most efficient methods for handling pattern-based events based on real-time data streams. Additionally, CEP[3] works with a predefined set of rules that are designed based on health parameters to aggregate and correlate to reach predictions.

However, rule design is a challenging task as it requires domain knowledge as well as careful consideration of techniques for predicting cardiovascular risk from a data stream. Building on the proposed system, the following points outline the proposed work's main contribution.

1. We propose a real-time model approach for cardiovascular disease prediction using Apache Spark and Kafka, with various cardio parameters.
2. We develop a predefined set of rules using fuzzy logic and a certified standard parameters for cardiovascular disease monitoring and prediction through the Siddhi CEP engine.
3. We propose a structured alert system for cardiovascular disease for enhancing more accurate prediction.
4. Validation of proposed method using synthetically generated data and how efficiently the disease can be diagnosed.
5. Analysis of event processing scenarios using time windows based intervals.
6. A real-time dashboard for analyzing CVDs prediction and monitoring changing parameters.

The remaining section of the research work follows as:- Section **II** discusses the background analysis and related work. Section **III** presents the proposed system and its working functionality. Section **IV** elaborates on this work's experimental details and results. Finally, Section **V** discusses conclusion, and future work.

## 2. Background analysis and related Work

In order to analyze the research gaps and how the proposed solution addresses them, it is important to discuss the existing work in the domain and its challenges before delving deeper into the sections.

### 2.1 Apache Spark and Kafka

Spark is one of the mostly used frameworks for simulating real time applications and supports a variety of workloads including batch processing, interactive queries, real-time streaming, and machine learning, all within a single framework. This eliminates the need for separate systems and simplifies development and maintenance. Its speed, ease of use, fault tolerance, and extensibility make it a popular choice for big data processing and analytics tasks. Spark can be directly integrated with Kafka for offset and processing data in real time[4][5]. Both Kafka and Spark support fault tolerance inherently by distributing data across multiple nodes and recovers in case of any failure.

The combination of Kafka and Spark Streaming is highly scalable. Kafka can handle large volumes of data by partitioning topics among multiple brokers in the cluster. Spark Streaming facilitates windowed computations, grouping data into intervals (e.g., every 10 seconds) and performing operations like aggregations on each window [6][7].

### 2.2 Complex Event Processing

CEP is a method used for analyzing high volumes of data that processes and analyzes real-time event streams and triggers responses based on patterns in high-speed information systems. Applications that need to discover complex patterns across several event streams in milliseconds need this technology. Financial, telecommunications, health care, transportation, and security businesses use CEP to make real-time decisions [8]. In CEP, patterns in these streams indicate simple patterns, which include a single occurrence that fits requirements, while complex patterns have non-obvious links between several events. Systems can react instantly to new data as they process these events in real time or near real time. CEP systems define patterns using rules or queries. These rules allow extensive temporal logic and event linkages in specialized CEP languages [9][10].

### 2.3 Fuzzy rules

Fuzzy logic systems use fuzzy rules as a key component to model approximate reasoning rather than fixed and exact reasoning. It is an extension of classical set theory; fuzzy logic introduces the concept of partial truth values between "completely true" and "completely false."

Fuzzy rules provide flexible thinking and decision-making in complex systems with ambiguity and vagueness, where binary logic fails. Its help systems behave more like humans, handling complexities more intuitively and adaptably. Fuzzy logic allows for the representation of vague or subjective concepts by assigning membership degrees to different categories or conditions [11].

In the medical field, it is common to encounter knowledge gaps. often remains questionable and uncertain. Fuzzy logic is a variant of multi-valued logic that goes beyond binary values. The true/false paradigm involves assigning truth values to variables between 0 and 1. This methodology aims to tackle imprecision and uncertainty by providing a mathematical solution. The foundation of fuzzy logic lies in the idea that human decision-making often relies on imprecise and non-numeric data. Fuzzy models, or fuzzy sets, are mathematical tools for capturing information vagueness and imprecision.

**2.4 Related work**

Hsu et al. [12] proposed a methodology for cardiovascular disease-related event detection using recurrent neural networks. The research work was carried out using a 2-year observation and 5-year prediction window. The experimental result shows that the proposed model achieved an average precision of 0.425 and an AUROC of 0.805.

Rumsfeld et al. [13] explored the challenges related to cardiovascular disease in the context of big data applications and emphasized the role of big data in improving the prediction model's accuracy.

Rahmani et al. [14] proposed an event-driven model for intelligent healthcare monitoring using CEP. The model uses real-time healthcare data for monitoring and data analysis. Additionally, the model is cost-effective and reliable enough for real-time healthcare applications.

Terrada et al. [22] proposed a fuzzy logic-based system designed to optimize patient diagnosis by integrating cardiovascular risk factors, organizing decision rules, and employing a fuzzification-defuzzification process to manage clinical data. The paper outlines the structure of this fuzzy cardiovascular diagnosis system and validates its performance using sensitivity and specificity measures, demonstrating the system's potential in improving diagnostic accuracy and minimizing errors.

Ma et al.[23] proposed a novel approach of prediction of coronary heart disease(CHD) for diabetic patients using an AI based model. They used statistics for the experiment to examine the distribution of four different types of features (basic demographic information, laboratory indicators, a medical exam, and a questionnaire) among comorbidities. Then, we tested how well three common machine learning methods could predict what would happen (extreme gradient boosting, random forest, and logistic regression). Table 2 describes the existing methodologies for the CEP in the health care domain, as well as its comparative analysis.

Table 2 Comparison of Existing Literature Review.

| Reference | Application | Research Focus | Methodology | Results |
|---|---|---|---|---|
| Naseri et al. (2021) [1] | Health Monitoring Systems | Intelligent rule extraction for health monitoring using CEP platforms | Developed a CEP-based platform utilizing rule-based learning (PART and JRip) for real-time health data processing | Achieved high accuracy in classifying health conditions using PART (98.61%) and JRip (97.32%) with fewer rules generated by JRip, leading to a more efficient CEP system. |
| Sun et al. (2019) [3] | Environmental Monitoring | Complex event processing for anomaly detection in environmental data. | A complex event processing (CEP) engine using Apache Kafka for real-time data processing and anomaly detection in geological carbon sequestration.. | Demonstrated effective real-time anomaly detection, enhancing intelligent monitoring systems with minimal coding required for domain users. |

| Hasan et al. (2020) [4] | Healthcare | Real-time health monitoring using machine learning. | Use of Spark streaming and online machine learning via a Spark MLlib algorithm to analyze data from wearable medical devices for health monitoring. | Achieved high accuracy rates (up to 98% in certain setups) in predicting health conditions in real-time, enabling effective monitoring and potential emergency alerts. |
|---|---|---|---|---|
| Kumar and Agarwal (2024) [8] | IoT applications | Rule-based complex event processing (CEP) for IoT | Comprehensive survey of CEP using rule-based algorithms, including event producers, preprocessing of events, robust rule implementation, and decision support | Presented a detailed classification of approaches and insights into designing robust rules, scalable event-driven architectures, and the challenges in rule-based CEP for streaming IoT data. |
| Naseri et al. (2022) [9] | Remote Health Monitoring | Adaptive user behavior modeling for health systems | Proposed an adaptive complex event processing (CEP) platform that utilizes rule-based learning and a personalized rule adaption method for remote health monitoring. | Showed significant improvements in accuracy (up to 15% for general adaptation and 3-6% for personalized rules) compared to non-adaptive systems, highlighting effectiveness in adapting to user-specific |

| Wang and Mendel (1992) [10] | Control Systems and Signal Processing | Generating fuzzy rules from numerical data for control systems | Developed a method to generate fuzzy rules from numerical data, integrating both numerical and linguistic information into a fuzzy rule base. | Demonstrated effective approximation of continuous functions and improved control system performance compared to other methods. |
|---|---|---|---|---|
| Hsu et al. (2022) [12] | Healthcare | Cardiovascular disease prediction using deep learning | Utilized long short-term memory (LSTM) neural networks to model multivariate sequential data for predicting cardiovascular disease events. | Demonstrated that LSTM outperformed traditional models in predicting cardiovascular events, with an AUROC of 0.801 and average precision of 0.425, highlighting the effectiveness of temporal models in clinical decision support. |
| Rahmani et al. (2021) [14] | Healthcare | Event-driven IoT architecture for reliable healthcare using complex event processing | Developed an IoT architecture integrated with a CEP platform, focusing on the health sector, to improve data analysis and decision-making based on real-time health data. | The architecture increased system reliability, reduced costs, and improved healthcare quality by enabling real-time data analysis and decision-making. |
| Šimšek et al. (2024)[15] | Air Quality Monitoring | AI-powered and fog-based predictive CEP system for air quality monitoring | Developed an AI and fog computing-based complex event processing system (CepAIr) using RNN, LSTM, CNN, and SVR models for air pollution prediction. | Demonstrated high success in predicting pollutant gas concentrations with deep learning models, especially CNN, while ensuring minimal network delay. |
| Chandra et al. (2023) [19] | Liver Diseases Prediction | Semantic Rule and Decision Tree (DT) rule based Liver Diseases Prediction | Developed a Basic Formal Ontology (BFO) based Liver Disease Ontology and semantic rules for batch processing and SPARQL query-based liver disease prediction and diagnosis. | The model is evaluated based on ontology metrics and event-based metrics, and the calculation of Precision, Recall, and F1 Score is performed based on Decision Trees (DT). |
| Li et al. (2023)[24] | Real-time health monitoring | Using deep learning for effective ECG signal classification. | Employed convolutional neural networks (CNNs) for feature extraction and classification of ECG signals. | Demonstrated high classification accuracy, providing a basis for reliable real-time cardiac monitoring systems. |

## 3. Materials and Methodologies

In this section, we introduce a layered model for intelligent cardiovascular disease prediction. The model integrates Apache Spark, Apache Kafka, and Siddhi framework, offering a comprehensive solution. We provide a detailed analysis of the proposed approach.

### 3.1 System Architecture of Fuzzy Rules based Intelligent Cardiovascular Disease Prediction

In IoT-based healthcare applications, rules are pivotal in identifying diseases based on early symptoms and recommending appropriate medical guidelines for treatment. However, managing such issues in real-time is complex due to changing parameters and the necessity of maintaining high standards for accurate disease prediction. Given this information, we propose an intelligent cardiovascular disease prediction system utilizing a fuzzy rule-based Complex Event Processing (CEP) approach. This system predicts diseases based on real-time healthcare parameters and designed rules. Figure 1 shows the architecture of the proposed system integrated with various frameworks.

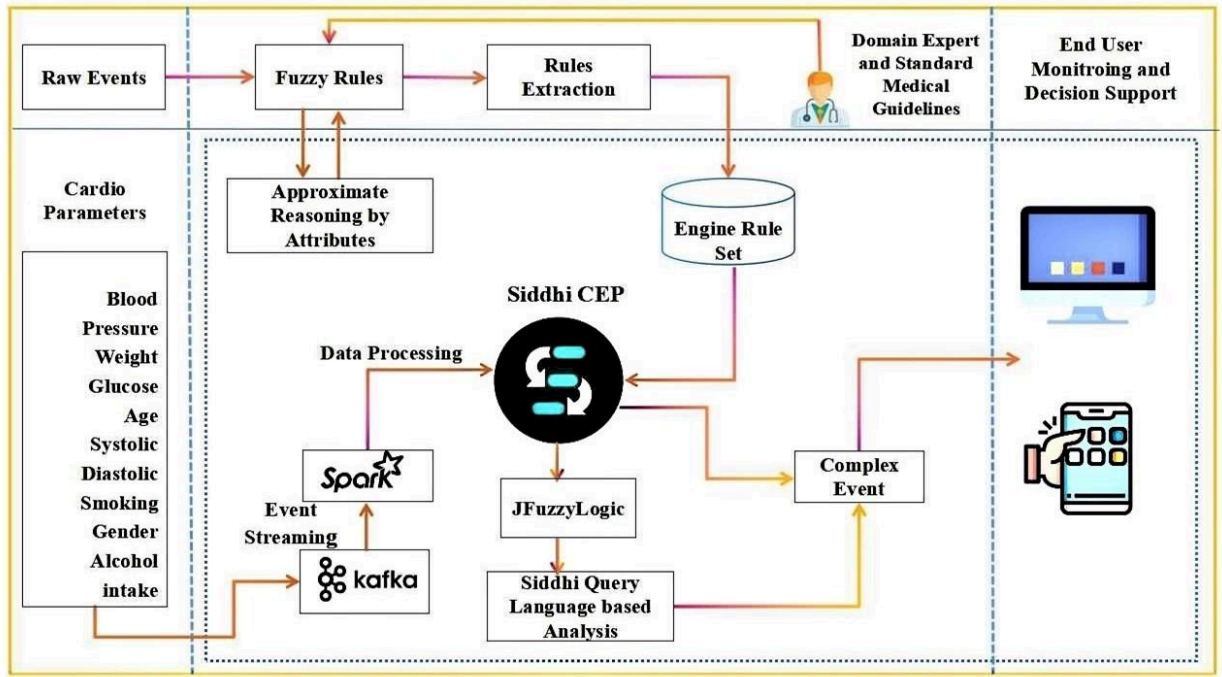

Fig.1. Proposed Architecture for Fuzzy rules based Intelligent Cardiovascular Disease Prediction

In the proposed system, there are three components: Event producer, Event processing, and Event consumers. The Event producer serves as the initial stage of this architecture, responsible for collecting cardiovascular parameters from sources, repositories, or raw data obtained from health sensors.

In our case, data has been collected from the Kaggle repository, where it is available as open source. Event Processing is the second phase of this architecture, using frameworks such as Apache Spark for handling real-time data, Kafka for managing large numbers of events, and Siddhi CEP for analyzing and correlating cardiovascular parameters in real-time. Fuzzy logic aids in designing approximate reasoning rules based on these parameters. At its core is the concept of a membership function, which determines the extent to which an input value belongs to a specific set or category.

Furthermore Siddhi CEP are used for correlating and analyzing the cardiovascular disease parameters from different sources based on fuzzy rules. The third component of the proposed architecture is an event consumer: a real-time dashboard developed to support the end user in better understanding and diagnosing CVDs.

**3.2 Apache Spark and Kafka Integration**

The integration of Kafka and Spark combines real-time data streaming and distributed processing. Kafka acts as a reliable data source, allowing Spark Streaming to consume data from Kafka topics. Spark processes the data using familiar APIs, and the results can be stored in various sinks. This powerful combination enables organizations to build robust, scalable, and real-time data pipelines for use cases like analytics and recommendations systems [16] [17] . Figure 2 illustrates the integrated approach of Kafka and Spark for stream processing.

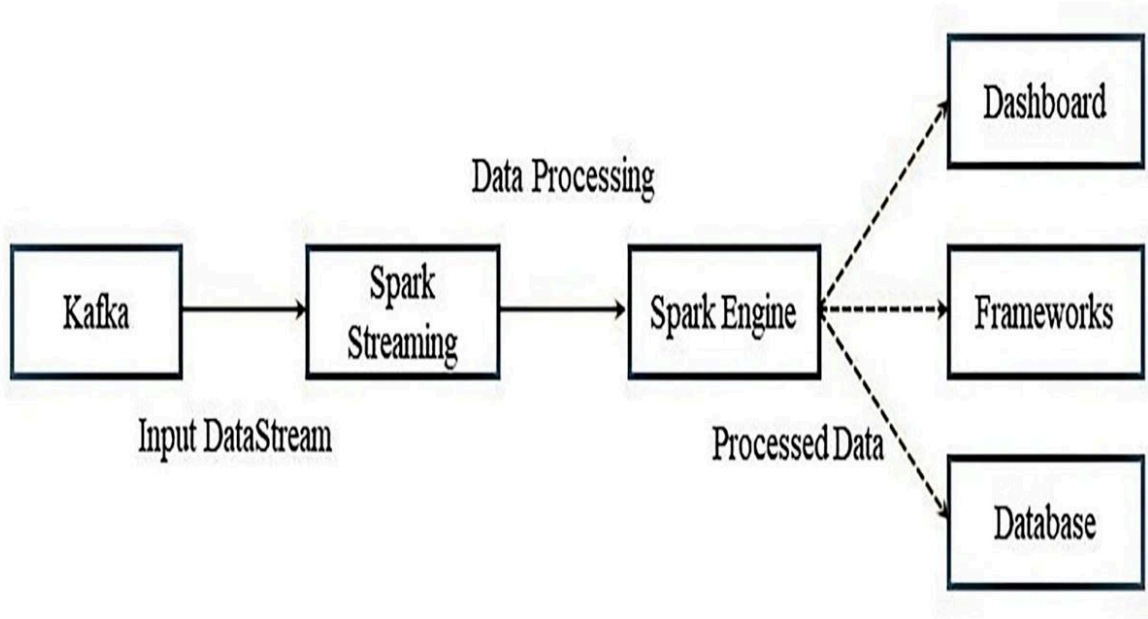

Fig 2. An integration of Kafka and Spark

**3.3 Siddhi Architecture**

Siddhi is an open-source, lightweight framework for real-time event processing. It handles large event volumes, extracting useful information from event streams. Commonly used for streaming analytics, rule-based decision-making, and adaptive systems, Siddhi supports query processing with user-defined rules. By correlating and analyzing events, it draws important conclusions that support decision-making for end-user applications[18]. In this work, the system takes cardiovascular parameters as input and predicts whether a patient has CVDs based on rules designed using medical standards and fuzzy logic. Figure 3 shows the working of Siddhi CEP engine with real time data streams.

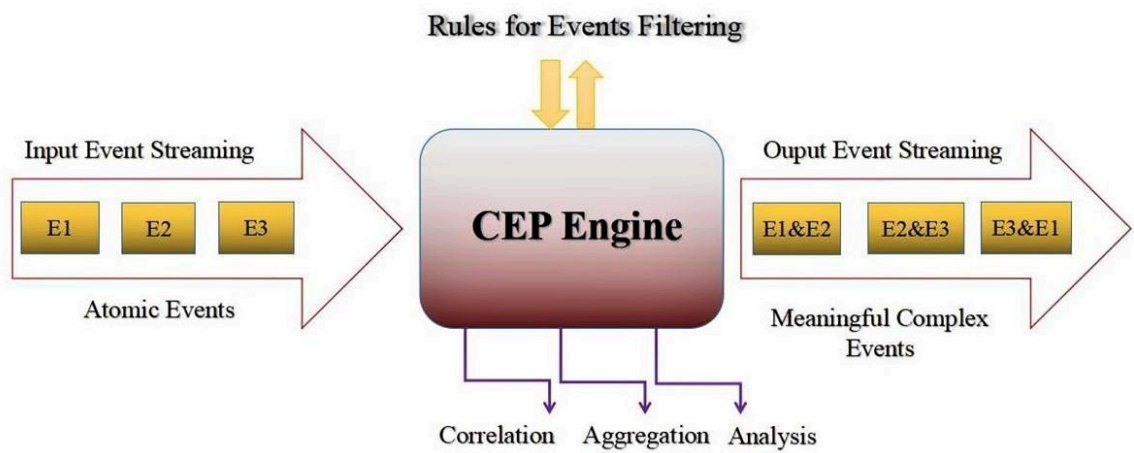

Fig. 3. Working of siddhi CEP engine

### 3.4 Working of Fuzzy Terminology

The term "fuzzy" means ambiguity or lack of clarity. In the medical field, knowledge often remains uncertain. Fuzzy Logic, a variant of multi-valued logic, goes beyond the binary true/false paradigm by assigning truth values to variables as real numbers between 0 and 1. This methodology provides a mathematical framework for handling imprecision and uncertainty in decision-making processes, based on the premise that human decision-making often relies on imprecise and non-numeric data. Fuzzy models, or fuzzy sets, capture vagueness and imprecision in information[19][20][21].

At the core of Fuzzy Logic is the concept of a membership function, which determines the extent to which an input value belongs to a specific set or category. This function maps input values to membership degrees ranging from 0 to 1. Fuzzy Logic operates through Fuzzy Rules, which are if-then statements connecting input and output variables in a fuzzy manner. The result is a fuzzy set with membership degrees for every potential output value. Different types of membership functions, such as Gaussian, triangular, trapezoidal, and sigmoidal, can represent a fuzzy variable[25][26][27][28][30]. Figure 4 illustrates the working of fuzzy logic and its membership functions.

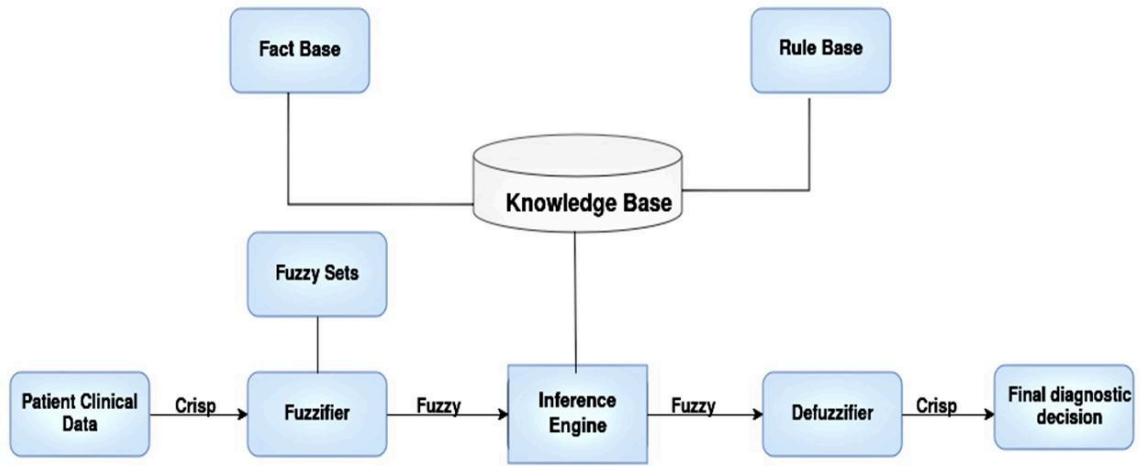

Fig 4. Working of fuzzy logic terminology

### 3.4.1 Membership Function

The input variables mentioned above are represented using triangular membership functions. A triangular membership function is defined by three parameters: a, b, and c, such that (a < b < c). These parameters form the three vertices of the triangular membership function. The mathematical definition of the triangular membership function is shown below.

$$\mu(x) = \begin{cases} 0, & \text{if } x \leq a \\ x-a/b-a, & \text{if } a \leq x \leq b \\ c-x/c-b, & \text{if } b \leq x \leq c \\ 0 & \text{if } x \geq c \end{cases} \quad (1)$$

A membership function in fuzzy logic defines how each point in the input space is mapped to a degree of membership between 0 and 1. It determines the extent to which an input value belongs to a specific fuzzy set or category. Fuzzification converts crisp input values into fuzzy values using membership functions. For example, if the age is 45, then the fuzzification process determines the category to which age belongs to the fuzzy sets as young, old and very old. Figure 5 depicts the membership function of age attribute and its value derived after fuzzification.

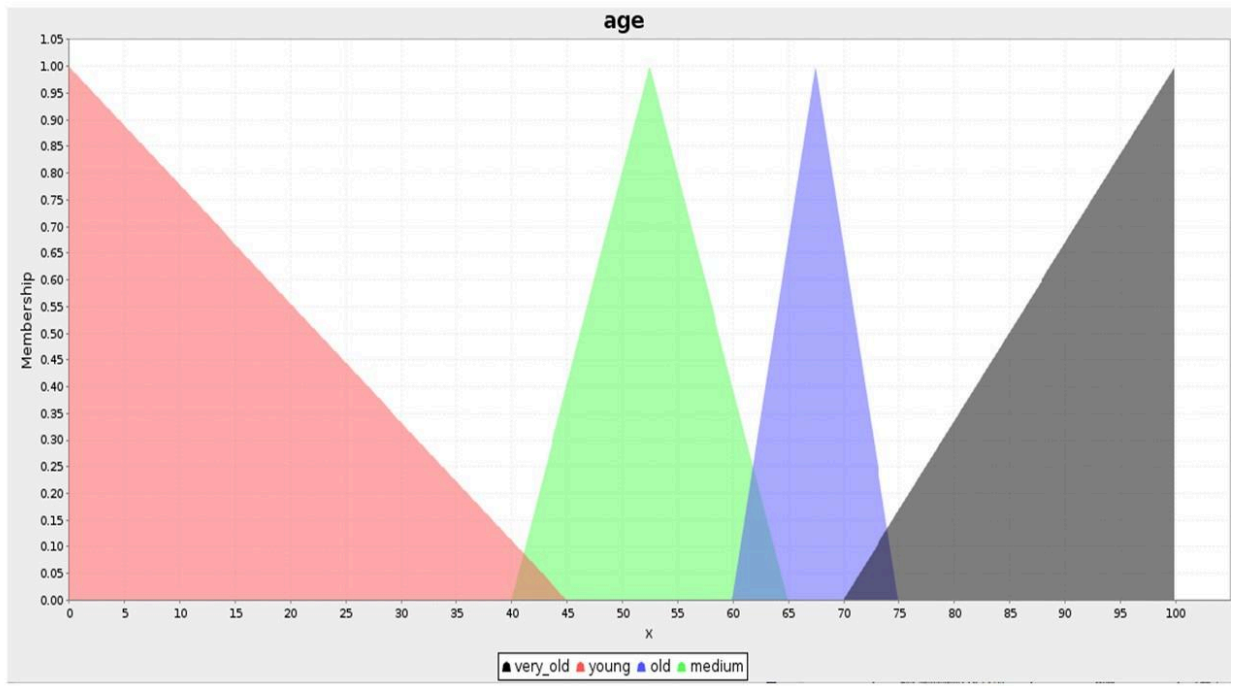

Fig 5. Membership function of age attribute.

**3.4.2 JFuzzyLogic**

The fuzzy logic in our research has been implemented using a Java library named JFuzzyLogic .It provides a framework for creating fuzzy logic systems and performing operations such as fuzzification, rule evaluation, and defuzzification. In JFuzzyLogic, FCL files are used to define fuzzy logic systems in a human-readable and easily understandable format. These files typically contain declarations of linguistic variables, membership functions, fuzzy rules, and other parameters necessary for fuzzy inference [29]. Figure 6 shows a sample of our FCL file where we have defined our membership functions and rule block.

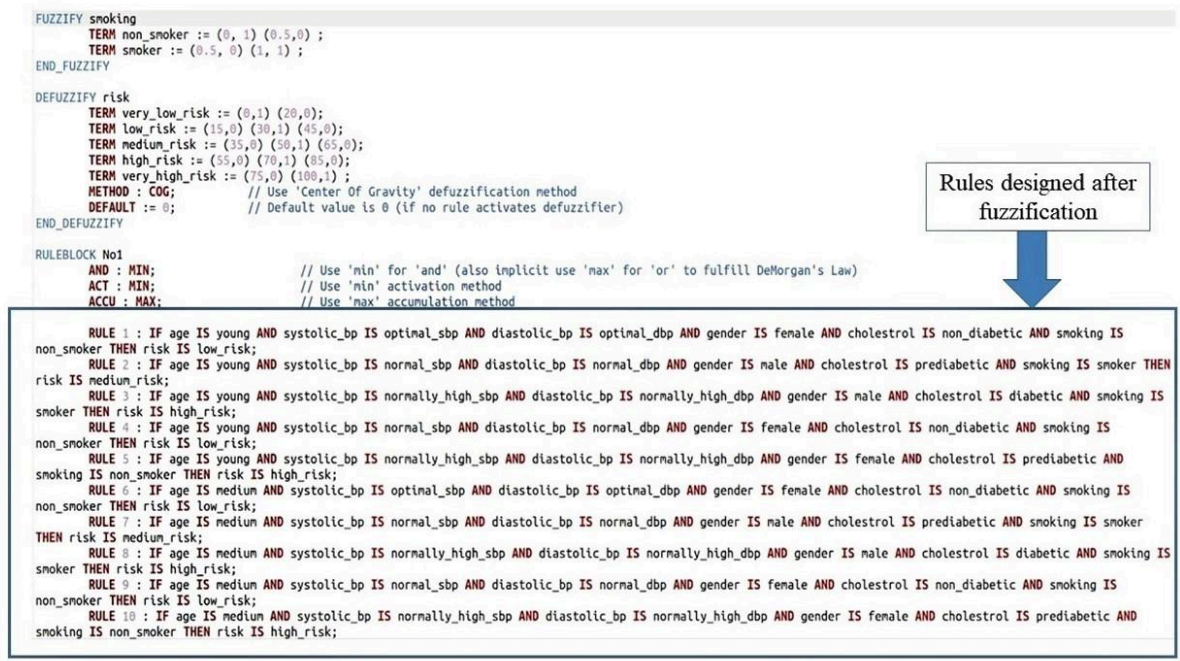

Fig.6. Working of Fuzzy Logic Terminology

## 3.5 Algorithm for rule development

**Input :** Blood Pressure, Weight, Glucose, Age, Systolic, Diastolic, Smoking, Gender, Alcohol intake

**Output :** Rule development for CEP based CVDs prediction.

**Step 1 :** Collect and preprocess data.

**Step 2 :** Take parameters and construct if then rules using medical guidelines.

**Step 3:** Modeling with the conventional fuzzy logic method.

**Step 4:** Apply approximate reasoning to fuzzify each attribute.

**Step 5:** Calculate membership functions to determine attribute values.

**Step 6:** Apply the fuzzy rules on CEP engine for CVDs prediction.

### 3.5.1 Standard Guidelines for CVDs

The rule base is a crucial component of our system. It maps input or computed truth values to desired output truth values using predefined rules. These rules are derived from cardiovascular disease symptoms and clinical guidelines from authorized organizations. The table shows the standard parameters for designing these rules[4]. Table 3 shows the standard guidelines used for developing the rules.

Table 3 Standard guidelines for CVDs prediction.

| Cardiovascular Risk Factor | Description of States | |
|---|---|---|
| **Age according to WHO** | **State** | **Age** |
| | Young | 45 |
| | Medium | 40-65 |
| | Old | 60-75 |
| | Very Old | >=70 |
| **SBP** | **SBP** | **mmHg** |
| | Optimal | <120 |
| | Normal | 120-129 |
| | Normally High | 130-139 |
| | Grade 1 (Light) | 140-159 |
| | Grade 2 (Moderate) | 160-179 |
| | Systolic Hypertension (Isolated) | >=140 |
| **DBP** | **DBP** | **mmHg** |
| | Optimal | <80 |
| | Normal | 80-84 |
| | Normally High | 85-89 |
| | Grade 1 (Light) | 90-99 |
| | Grade 2 (Moderate) | 100-109 |
| | Grade 3 (Severe) | >=110 |
| | Diastolic Hypertension (Isolated) | <90 |
| **Gender** | **Gender** | **Value** |
| | Female | 1 |
| | Male | 2 |
| **Smoking** | **Smoking** | **Value** |
| | Non-Smoker | 0 |
| | Smoker | 1 |
| **Diabetes** | **Diabetes** | **(g/l)** |
| | Non-diabetic | 0.8-1.0 |
| | Prediabetic | 1.01-1.24 |
| | Diabetic | >=1.26 |
| **Cholesterol** | **Cholesterol** | **mmol/l** |
| | Normal | <5.2 |
| | Medium | 5.2-6.1 |

---

[4] https://www.acpicr.com/data/Page_Downloads/ACPICR2023StandardsReaderlayout.pdf

| | | High | >=6.2 |
|---|---|---|---|

### 3.5.2 Rule Designed using Clinical Guidelines.

Based on medical guidelines and fuzzy logic, we designed around 30 rules for filtering CVD parameters. Table 4 shows some of these rules for CEP-based cardiovascular disease prediction.

Table 4 Fuzzy logic based rules development.

| S.No | If | Then |
|---|---|---|
| RULE 1 | IF age IS young AND systolic_bp IS optimal_sbp AND diastolic_bp IS optimal_dbp AND gender IS female AND cholestrol IS non_diabetic AND smoking IS non_smoker | Risk IS low_risk; |
| RULE 2 | IF age IS young AND systolic_bp IS normal_sbp AND diastolic_bp IS normal_dbp AND gender IS male AND cholestrol IS prediabetic AND smoking IS smoker THEN | Risk IS medium_risk; |
| RULE 3 | IF age IS young AND systolic_bp IS normally_high_sbp AND diastolic_bp IS normally_high_dbp AND gender IS male AND cholestrol IS diabetic AND smoking IS smoker | Risk IS high_risk; |
| RULE 4 | IF age IS young AND systolic_bp IS normal_sbp AND diastolic_bp IS normal_dbp AND gender IS female AND cholestrol IS non_diabetic AND smoking IS non_smoker | Risk IS low_risk; |
| RULE 5 | IF age IS young AND systolic_bp IS normally_high_sbp AND diastolic_bp IS normally_high_dbp AND gender IS female AND cholestrol IS prediabetic AND smoking IS non_smoker | Risk IS high_risk; |
| RULE 6 | IF age IS medium AND systolic_bp IS optimal_sbp AND diastolic_bp IS optimal_dbp AND gender IS female AND cholestrol IS non_diabetic AND smoking IS non_smoker THEN risk IS low_risk; | Risk IS medium_risk; |
| RULE 7 | IF age IS medium AND systolic_bp IS normal_sbp AND diastolic_bp IS normal_dbp AND gender IS male AND cholestrol IS prediabetic AND smoking IS smoker | Risk IS medium_risk |

| RULE 8 | IF age IS medium AND systolic_bp IS normally_high_sbp AND diastolic_bp IS normally_high_dbp AND gender IS male AND cholestrol IS diabetic AND smoking IS smoker | Risk IS high_risk; |
|---|---|---|
| RULE 9 | IF age IS medium AND systolic_bp IS normal_sbp AND diastolic_bp IS normal_dbp AND gender IS female AND cholestrol IS non_diabetic AND smoking IS non_smoker | Risk IS low_risk; |
| RULE 10 | IF age IS medium AND systolic_bp IS normally_high_sbp AND diastolic_bp IS normally_high_dbp AND gender IS female AND cholestrol IS prediabetic AND smoking IS non_smoker | Risk IS high_risk; |

**Experiment Details and Results**

This section presents the experiment details and results obtained by rigorously filtering important parameters from the real-time data stream. To perform the experiment, we used a PC with 16.0 GB of RAM, a 64-bit OS, an x64 processor, Windows 10 Pro Edition, and version 21H2.

**4.1 Dataset Description**

We used an open-source cardiovascular disease dataset from the Kaggle repository, containing 70,000 samples and 11 cardio parameters. These include age, height, weight, gender, systolic blood pressure, diastolic blood pressure, cholesterol, glucose, smoking status, alcohol intake, physical activity, and the presence or absence of cardiovascular disease as the target value. The dataset contains no missing or null values, so data preprocessing is not required.

**4.2 Time Window based Execution Analysis**

Window interval-based event analysis was performed using a 5-second interval to analyze real-time event processing. Figure 7 demonstrates the model's efficient processing of cardiovascular parameters within small windows. As the window length increases, the number of events also increases, aiding in efficient disease diagnosis based on conditional rules. We have considered five rules at a time for processing cardio-based events.

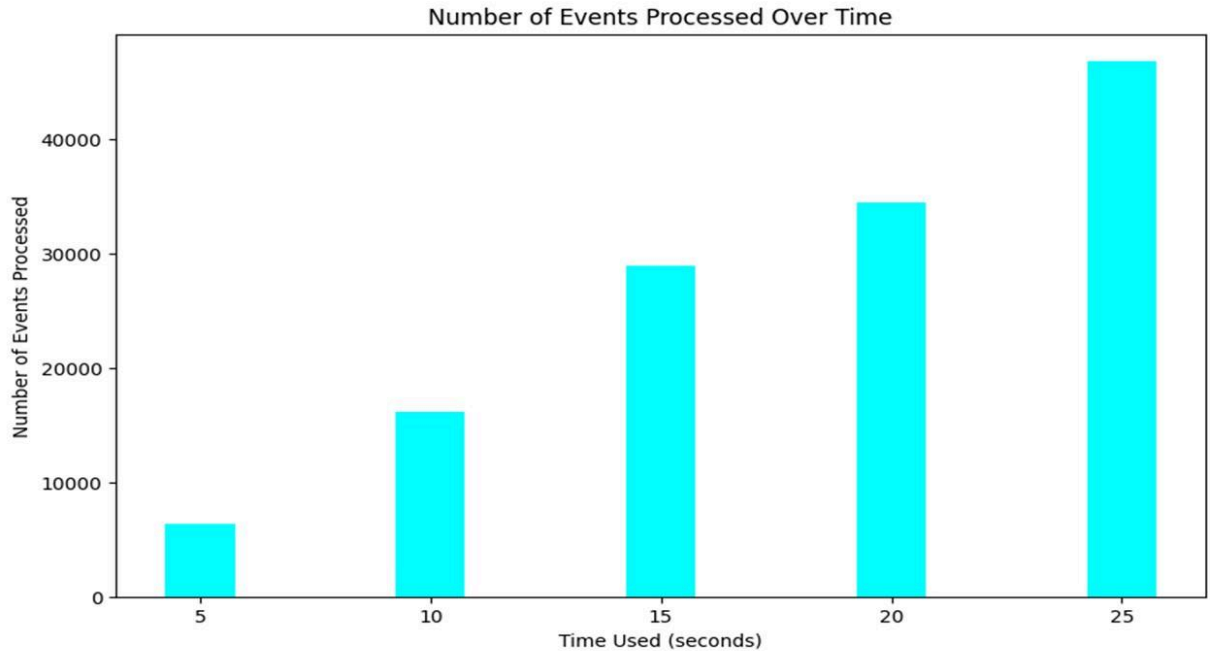

Fig.7 Number of event processed using time length window

Table 5. Comparative analysis of rules processed across different time windows and events.

| Time Used (s) | Events Processed (5 Rules) | Events Processed (10 Rules) | Events Processed (15 Rules) | Events Processed (20 Rules) | Events Processed (25 Rules) |
|---|---|---|---|---|---|
| 5 | 6397 | 4572 | 4392 | 4026 | 3528 |
| 10 | 16103 | 10077 | 9197 | 9255 | 8479 |
| 15 | 28884 | 22505 | 15754 | 13906 | 11790 |
| 20 | 34442 | 30429 | 20957 | 18124 | 15096 |
| 25 | 46837 | 36123 | 26671 | 22910 | 18976 |

Table 5 demonstrates the impact of the number of rules on the event processing capacity over various time intervals. As the number of rules increases, the number of events processed decreases across all time intervals.

### 4.3 Rule Deployment Time for CEP Engine

In this section we consider deploying CEP rules to CEP engines to determine how efficiently rules are being processed by a stream of events[31][32]. In this scenario we have considered different counts of rules. Figure 8 shows the time it takes to deploy a rule based on the number of events processed [33]. Additionally, the execution time increases with the number of rules and events, reaching up to 25 seconds at maximum load.

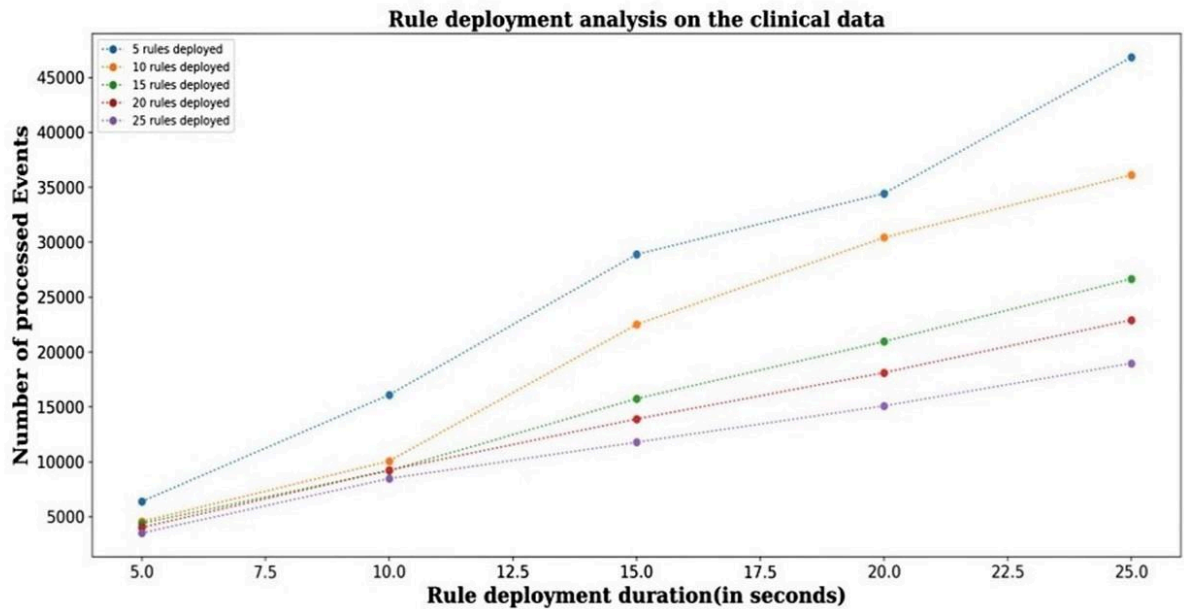

Fig.8. Rule deployment time to CEP engine

### 4.4 Average Number of Event Detection for Time Window Length

The average event detection window length significantly impacts accuracy. Shorter windows enhance sensitivity to rapid changes but may introduce noise. This analysis aims to detect changes in CVDs parameters and take prompt action based on specified rules. Figure 9 shows that the number of events increases with longer time windows.

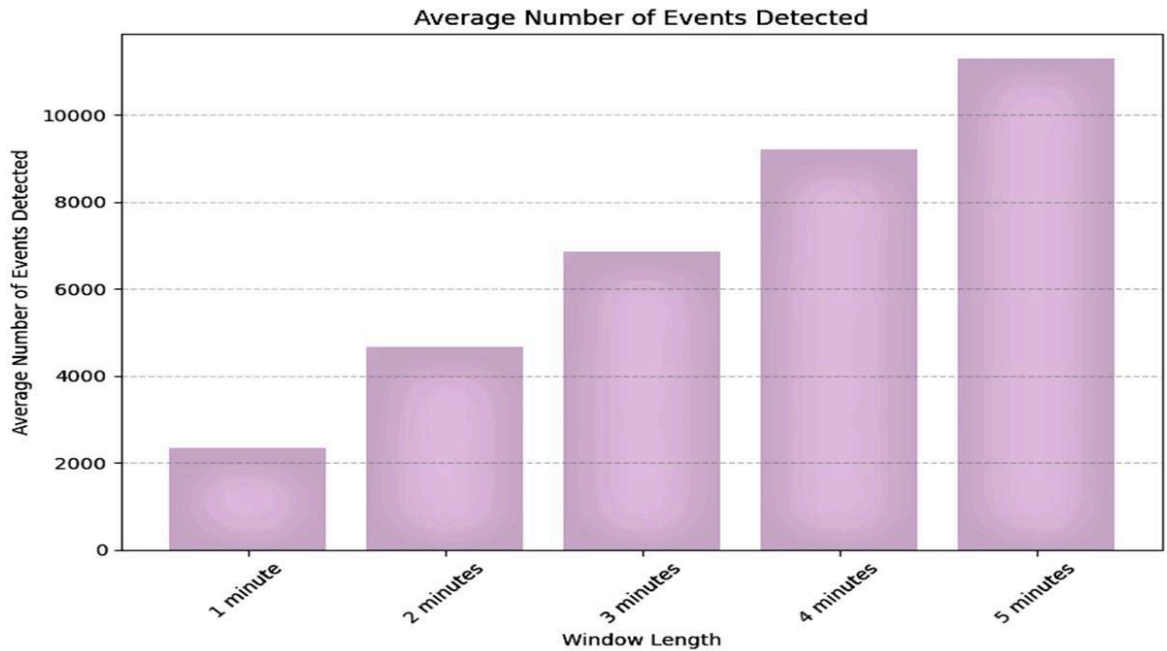

Fig.9. Time window length based event detection

**4.5 Validation of Proposed Approach**

To validate our approach, we used synthetic data to test the efficiency of our CEP-based model in predicting CVDs. We generated 1000 samples using Python's Numpy library, applying parameters to validate the model on unseen data. The model predicts the presence or absence of CVDs and, if present, the disease's severity using designed rules.

Figure 10 shows that less than 20% of samples were categorized as very low risk, 15-45% as low risk, 35-65% as medium risk, 55-85% as high risk, and 75% as very high risk. This correct categorization demonstrates the effectiveness of the designed rules and model. Table 6 illustrates the categorization of the synthetic data into different levels of CVDs.

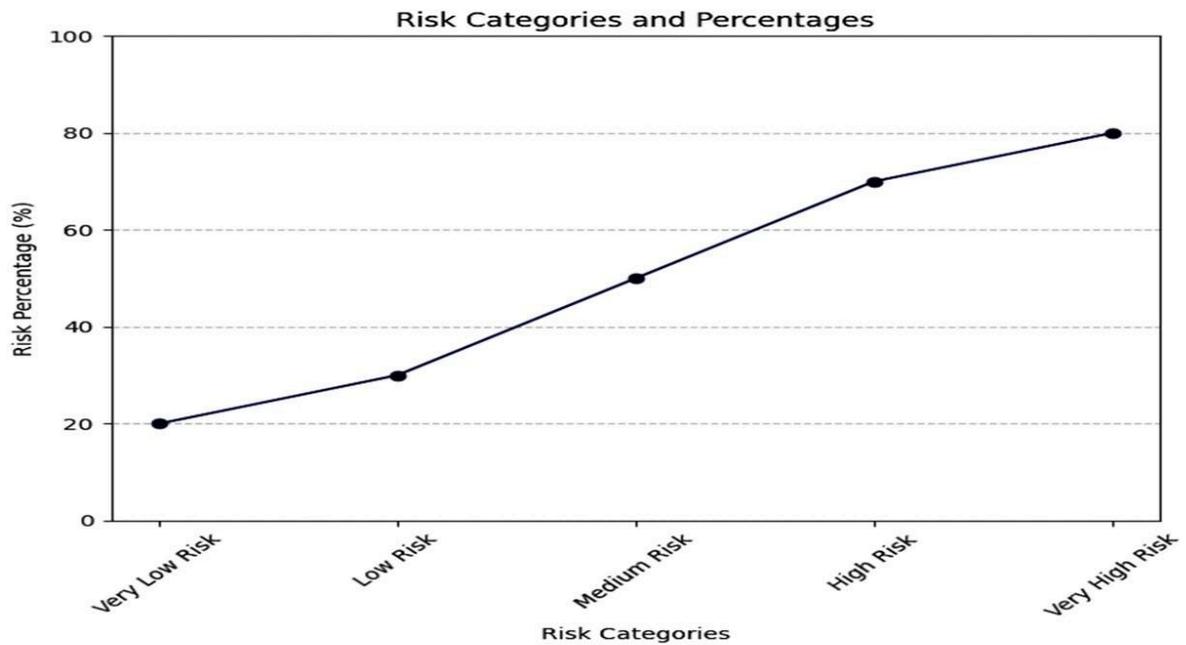

Fig.10. Risk percentage classification of disease prediction.

Table 6. CVDs classification based on input data.

| Risk Category | Number of Samples |
|---|---|
| Very Low Risk | 237 |
| Low Risk | 226 |
| Medium Risk | 106 |
| High Risk | 156 |
| Very High Risk | 275 |

We developed a real-time dashboard for predicting CVDs. This dashboard acts as an event consumer, integrating sensor data processed by our CEP system. It provides healthcare professionals and patients with immediate insights into cardiovascular health by visualizing key metrics and fluctuations. The dashboard also offers risk assessments and severity categorizations based on predefined rules, enabling timely interventions and informed decision-making to enhance patient care.

Additionally, organizations can use the dashboard to support their decisions, customized to their specific requirements. Figure 11 illustrates a real-time dashboard that predicts CVDs based on cardio parameters, and it reflects changes in parameters efficiently.

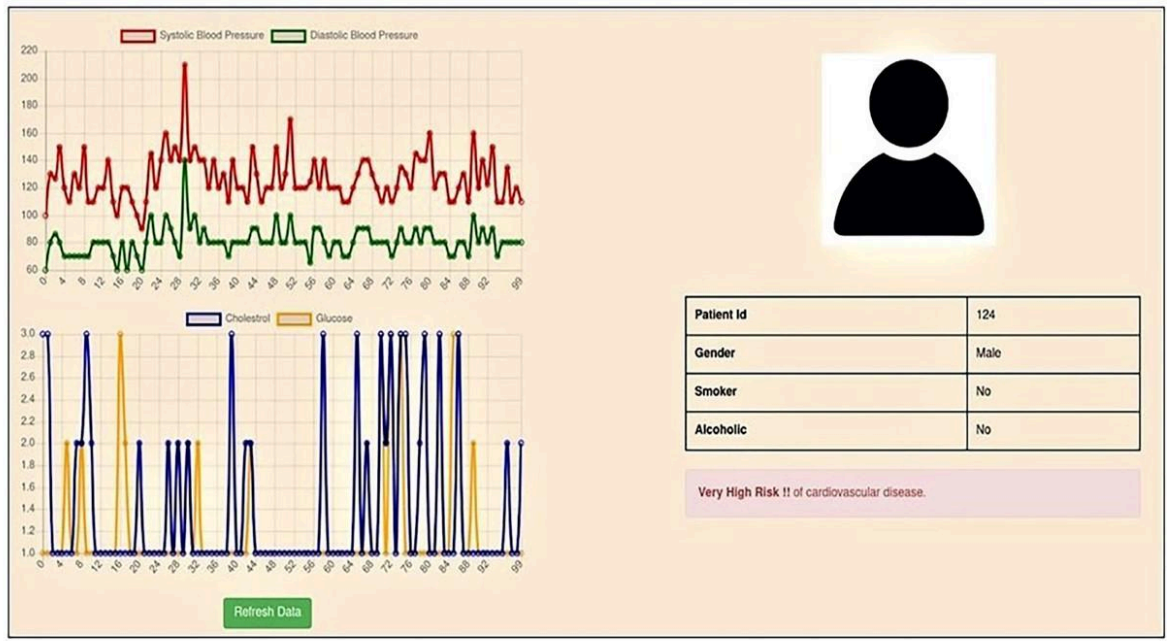

Fig.11. Real Time dashboard for CVDs prediction.

**Conclusion and Future work**

In this paper, we propose a fuzzy rule-based intelligent CVDs prediction model using CEP. Our real-time approach offers scalable and accurate predictions by employing rule-based parameters to simulate the model effectively. We integrated Apache Kafka and Apache Spark to handle large volumes of cardio parameter-based events in real-time scenarios. Additionally, we used a CEP engine to analyze these events, developing rules with fuzzy logic and standard medical guidelines for CVDs prediction. By processing a large number of events within specific time windows, we identified useful patterns and efficiently monitored input parameter fluctuations for accurate decision support.

We validated our model using synthetic data (1000 samples), categorizing each attribute into "Very Low Risk, Low Risk, Medium Risk, High Risk, and Very High Risk." The validation results showed that 20% of samples were categorized as very low risk, 15–45% as low risk, 35–65% as medium risk, 55–85% as high risk, and 75% as very high risk.

For future studies, we aim to enhance our research by extending it to a distributed CEP environment for fault tolerance and efficient event timestamping in CVD prediction. We also plan to optimize fuzzy logic, include more medical parameters, and integrate Semantic Web technology to improve data interoperability and enable advanced semantic reasoning for more accurate predictions, while enhancing the correlation capabilities of the CEP system.

**Acknowledgement**

We are thankful for Big Data Analytics Lab and Support Development Center TEAL 2.O at IIIT Allahabad for providing necessary resources for conducting this research.

**Declaration**

**Conflict of interest**: There is no conflict of interest with respect to the work reported in this paper.

**Financial assistance**: This research work received no funding.

**Availability of data and material**: The data and material is available on reasonable request.

**Authorship credit statement: Shashi shekhar kumar-** Formal Analysis, Conceptualization, Drafting original manuscript. **Anurag harsh-** Coding and implementation. **Ritesh chandra-** Reviewing, Conceptualization and editing original manuscript**. Sonali Agarwal-** Supervision throughout the work.